\documentclass[review,9pt]{elsarticle}

\usepackage{hyperref}
\usepackage{multirow}
\usepackage{color}
\usepackage{subfig}
\usepackage{soul}
\usepackage{breqn}
\usepackage[table,xcdraw]{xcolor}
\usepackage{stackengine}
\usepackage{tabularx}
\usepackage{todonotes}
\usepackage{booktabs}
\usepackage{xcolor}
\usepackage{soul}
\usepackage{todonotes}
\usepackage{amsmath,amsthm,amssymb}
\usepackage{graphicx}
\usepackage{adjustbox}
\usepackage{rotating}
\usepackage{hhline}
\usepackage{verbatim} 

\journal{Journal of Pattern Recognition Letters}

\begin{document}

\begin{frontmatter}

\title{Bi-Discriminator GAN For Tabular Data Synthesis\footnote{{Supplementary materials and source codes are available at} \href{https://github.com/EsmaeilpourMohammad/BCT-GAN.git}{this github-Repo.}}}

 \author{Mohammad Esmaeilpour$^{\mathsection \dagger}$\corref{}}
 \ead{mohammad.esmaeilpour.1@ens.etsmtl.ca} 
 \author{Nourhene Chaalia$^\dagger$}
 \ead{nourhene.chaalia@desjardins.com}
 \author{Adel Abusitta$^\ddagger$}
 \ead{adel.abusitta@polymtl.ca}
 \author{François-Xavier Devailly$^\dagger$}
 \ead{francois-xavier.a.devailly@desjardins.com}
 \author{Wissem~Maazoun$^\dagger$}
 \ead{wissem.maazoun@desjardins.com}
 \author{Patrick Cardinal$^\mathsection$}
 \ead{patrick.cardinal@etsmtl.ca} 
 \address{$^{\mathsection}$\'{E}cole de Technologie Sup\'{e}rieure (\'{E}TS), Universit\'{e} du Qu\'{e}bec and \\ IVADO Institution, Montr\'{e}al, Qu\'{e}bec, Canada\\
 $^{\dagger}$The Fédération des Caisses Desjardins du Québec\\
 $^{\ddagger}$\'{E}cole Polytechnique de Montr\'{e}al and McGill University\\
1100 Notre-Dame W, Montr\'{e}al, H3C 1K3, Qu\'{e}bec, Canada}




\begin{abstract}
This paper introduces a bi-discriminator GAN for synthesizing tabular datasets containing continuous, binary, and discrete columns. Our proposed approach employs an adapted preprocessing scheme and a novel conditional term using the $\chi^{2}_{\beta}$ distribution for the generator network to more effectively capture the input sample distributions. Additionally, we implement straightforward yet effective architectures for discriminator networks aiming at providing more discriminative gradient information to the generator. Our experimental results on four benchmarking public datasets corroborates the superior performance of our GAN both in terms of likelihood fitness metric and machine learning efficacy. 
\end{abstract}

\begin{keyword}
Generative adversarial network (GAN) , tabular data synthesis, bi-discriminator GAN, conditional generator, variational Gaussian mixture model (VGM). 
\end{keyword}

\end{frontmatter}


\section{Introduction}
\label{sec:intro}
Tabular data is among the most common modalities which has been widely used for maintaining massive databases of financial institutions, insurance corporations, networking companies, healthcare industries, etc. \cite{even2007economics,shwartz2021tabular,clements2020sequential,buczak2015survey,ulmer2020trust}. These databases include immense combination of personal, confidential, and general records for every customer, client, and patient in different formats (e.g., continuous and discrete data types). Semantic patterns derivable in such records efficiently contribute to extract meaningful information for the benefit of companies in various aspects such as large-scale decision-making \cite{xu2020synthesizing}, risk management \cite{aven2010risk}, long-term investment \cite{kornfeld1998automatically}, fraud or unusual activity detection \cite{cartella2021adversarial}, etc. However, exploiting these patterns is a challenging task since tabular datasets are heterogeneous \cite{sheth1990federated,wang1990polygen} and they contain sparse representations of discrete and continuous records with low correlation compared to homogeneous datasets (e.g., speech, environmental audio, image, etc.) \cite{borisov2021deep}. Unfortunately, extracting semantic relational patterns from heterogeneous datasets requires implementing costly data-driven algorithms \cite{loorak2016exploring,khan2020toward}.        

During the last decade and especially after the proliferation of deep learning (DL) algorithms, various cutting-edge approaches have been introduced for processing tabular datasets in different frameworks \cite{socher2012deep,traquair2019deep,gorishniy2021revisiting}, particularly for synthesis purposes \cite{bourou2021review}. Presumably, this is due to two major applications. Firstly, complex DL algorithms configured in the generative adversarial network (GAN) \cite{goodfellow2014generative} synthesis platforms can be used for augmenting sparse datasets with low cardinality and poor sample quality \cite{shanmugam2020and}. Often, this data augmentation procedure effectively triggers the semantic pattern extraction operations. Secondly, GAN-based synthesis approaches yield models capable of generating new records similar (and non-identical) to the ground-truth samples available in the original databases. The synthesized records can be used for development purposes such as extracting relational patterns \cite{tsechansky1999mining} without publicly releasing the original dataset. This efficiently contributes to protect the privacy of people and clients whose their information is stored in the tabular datasets of companies. Our focus in this paper is on the latter application since it has been among the most demanding appeals of some large-scale financial institutions towards avoiding data leakage \cite{shabtai2012survey,alneyadi2016survey}. Briefly, we make the following contributions in this paper:
\begin{enumerate}[(i)]
\item developing a novel data preprocessing scheme and defining a new conditional term (vector) for the generator network configured in a GAN synthesis setup,
\item implementing a bi-discriminator GAN for providing more gradient information to the generator network in order to improve its performance in runtime,
\item designing straightforward architectures for generator and discriminator networks.
\end{enumerate}
The organization of this paper is as the following. Section~\ref{sec:background} provides a summary of synthesis approaches based on the state-of-the-art GANs for tabular datasets. In Section~\ref{sec:proposed}, we explain the details of our proposed synthesis approach and finally in Section~\ref{sec:experiment}, we report and analyze our conducted experiments on four benchmarking databases.

\section{Background: Tabular Data Synthesis}
\label{sec:background}
Over the past years, variational autoencoder (VAE\cite{kingma2013auto}) and GAN frameworks have been recognized as the state-of-the-art approaches for data fusion, particularly in the context of tabular data synthesis \cite{xu2020synthesizing}. Fundamentally, these two frameworks are similar to the baseline classical Bayesian network (CLBN) \cite{chow1968approximating} and its variants such as private Bayesian network (PrivBN) \cite{zhang2017privbayes}. Thus far, many modern forms of these generative models (e.g., \cite{ma2020vaem,park2018data}) have been introduced and practically implemented for real-life applications. However, they both suffer from some major technical limitations and instability side-effects \cite{genevay2017gan}. In terms of comparison, there are some debates on their relative performance over another \cite{mi2018probe}, nevertheless we do not address them herein since they are out of the scope of the current work.

In this section, we briefly review the background of tabular data synthesis with a focus on GANs since we are mostly interested in developing generative models capable of synthesizing new records (a collection of fields) directly from random latent distributions \cite{goodfellow2014generative}. This is one of the pivotal characteristics of GANs which normally provides a wider synthesis domain and yields a more comprehensive generative model \cite{feizi2017understanding}. In general, a standard GAN configuration (i.e., vanilla GAN) employs two DL architectures organized in a mini-max setup as the following \cite{goodfellow2016deep}.
\begin{equation}
    \min_{G} \max_{D} \mathbb{E}_{\mathbf{x}\sim p_{r}}\left [ \log D(\mathbf{x}) \right ]+
\mathbb{E}_{\mathbf{z}\sim p_{z}}\left [ \log \left ( 1-D(G(\mathbf{z})) \right ) \right ]
\label{eq:gan}
\end{equation}
\noindent where the aforementioned architectures are denoted by $G(\cdot)$ and $D(\cdot)$ which represent the generator and discriminator networks, respectively. For $\mathbf{z}_{i} \in \mathbb{R}^{d_{z}}$ with dimension $d_{z}$ drawn from the multivariate random distribution $p_{z}$, the generator network synthesizes a record similar to the ground-truth sample indicated by $\mathbf{x}$. The control of this synthesis process is on the discriminator which samples from the real distribution $p_{r}$. Specifically, this process binds $D(\cdot)$ to measure and tune the convergence of the generator's distribution ($p_{g}$) to $p_{r}$ in an iterative pipeline.   

There are several outstanding variants for Eq.~\ref{eq:gan} employing different DL architectures, optimization properties, and loss functions. For instance, MedGAN \cite{choi2017generating} is among the premier generative models developed for synthesizing tabular datasets particularly for multi-label discrete records. This model exploits an autoencoder on top of the generator network to learn salient features of the continuous medical records. Recently, an extension for MedGAN has been released which imposes a regularization scheme on the generator using an adversarially training-based autoencoding policy \cite{camino2018generating}. Regarding the conducted experiments and reported results, although this scheme improves the performance of the generator for multi-categorical continuous records, it increases the chance of training instability and mode collapse.

A simpler yet effective introduced approach for synthesizing highly heterogeneous tabular datasets is Table-GAN \cite{park2018data} which is inspired from the conventional deep convolutional GAN configuration \cite{radford2015unsupervised}. Unlike MedGAN, it can be adapted for any tabular datasets with wide ranges of multi-categorical records. This generalizability mainly originates from the distinctive loss function of Table-GAN as follows.
\begin{equation}
    L_{G}(\cdot) = \mathbb{E}  \Big[ \left | \ell_{app}(G(\mathbf{z}_{i}))-C(\ell_{rm} (G(\mathbf{z}_{i}))) \right |  \Big]_{z_{i}\sim p_{z}(z)}
    \label{eq:tableGAN}
\end{equation}
\noindent where $\ell_{app}(\cdot)$ and $\ell_{rm}(\cdot)$ denote the analytical appending and removing functions for the label attributes (fields) of each column, respectively. Additionally, $C(\cdot)$ retrieves the output logit (predicted column label) of the generator network. In a nutshell, $L_{G}(\cdot)$ synchronizes the output logits of the generator to smoothly match the order of the original records and consequently improves the accuracy of the entire model. However, it might be computationally prohibitive for tabular datasets with numerous columns (features).

Another straightforward approach for generating complex combinations of continuous and discrete records without employing such a costly loss function in Eq.~\ref{eq:tableGAN} is introduced by Mottini {\it et al.}~\cite{mottini2018airline}. They propose a noble similarity measure using the Cram\'{e}r integral probablity metric (IPM \cite{mroueh2017sobolev}) \cite{bellemare2017cramer} in order to minimize the distribution discrepancies among original and synthesized samples. Following this technique, they incorporate a decomposition policy for sampling $\mathbf{z}_{i}$ as:
\begin{equation}
     \delta (\mathbf{z}_{i}) \doteq  \mathrm{KL}_{div} \Big[ p_{r} \left (\rho_{1} |\mathbf{z}_{i} \right)\parallel \rho_{2}  \Big], \quad \rho_{1},\rho_{2} \sim \mathcal{B}\Big(\frac{1}{2}\Big)
\end{equation}
\noindent where $\mathrm{KL}_{div}$ refers to the statistical Kullback-Leibler divergence, $\mathcal{B}$ indicates the Bernoulli's distribution, and $ \delta (\mathbf{z}_{i})$ is the normalized sampled vector for further reducing potential local discrepancy among the synthesized records. Mottini's GAN (MT-GAN) has been successfully evaluated on public tabular databases and it has been relatively outperformed other GANs with standard $\varphi$-divergence IPMs such as vanilla GAN~\cite{goodfellow2014generative}. 

ITS-GAN which stands for the incomplete table synthesis GAN \cite{chen2019faketables} is another reliable cutting-edge variant for Eq.~\ref{eq:gan}. One of the major novelties in this approach is the possibility of making a trade-off between accuracy and generalizability for the generator network. Technically, ITS-GAN exploits a massive combination of functional dependencies (FD: similar to the critic functions \cite{muller1997integral} in the sequence-to-sequence GAN platforms \cite{mroueh2017fisher}) for the continuous records using an independent pretrained autoencoder. However, this setting is primarily designed for small-scale datasets with relatively high correlation such as US Census \cite{kohavi1996scaling}. 

The class-conditional tabular GAN (CT-GAN) \cite{xu2019modeling} is a more sophisticated approach compared to ITS-GAN. For improving computational complexity it does not employ any costly FD modeling procedure. However, it implements a conditional generator (i.e., $G(\cdot)$ receives partial information of the columns in addition to $\mathbf{z}_{i}$) with mask vectors sampled form the original records. According to the published experimental results, CT-GAN confidently outperforms other advanced generative models for the public tabular datasets. In the following section, we extend this generative model to a bi-discriminator configuration with a novel definition for the conditional term and employing fully convolutional architectures. Finally in Section~\ref{sec:experiment}, we experimentally prove the superior performance of our proposed synthesis GAN on four popular benchmarking datasets.

\section{Proposed Approach: Bi-Discriminator class-conditional tabular GAN (BCT-GAN)}
\label{sec:proposed}
\begin{figure*}[t]
  \centering
  \includegraphics[width=\textwidth]{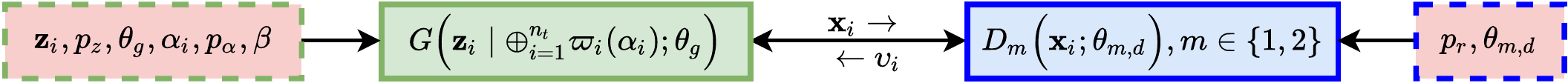}
  \caption{Overview of the proposed BCT-GAN. Herein, the random input vector is denoted by the multivariate $\mathbf{z}_{i} \in p_{z} \sim \mathcal{N}(0,I)$ and $\mathbf{x}_{i}$ represents the synthetic record (output). Additionally, $\theta_{g}$ and $\theta_{m,d}$ refer to the  weight vectors of the generator and $m$ independent discriminators, respectively. Moreover, $\upsilon_{i}$ is the returned logit vector from $D_{m}(\cdot)$ to $G(\cdot)$ for updating $\theta_{g}$. In the green rectangle, $\alpha_{i} \sim p_{\alpha}$ and $\varpi_{i}(\cdot)$ with parameter $\beta$ indicate the conditioning parameter (degree of freedom) and the condition vector for $G(\cdot)$, respectively. Rectangles depicted with green and blue dashed lines represent inputs to the networks.}
  \label{fig:overview-ECT}
  \vspace{-15pt}
\end{figure*}

Our proposed tabular synthesis approach is based on the CT-GAN \cite{xu2019modeling}, however with three major improvements in normalizing categorical records (preprocessing), defining the conditional term for the generator, and designing the architecture of the generator and two discriminator networks. The motivation behind employing double discriminators in our synthesis setup is the possibility of gaining more gradient information for the benefit of the generator during training. However, we do not provide the gradient analysis in this paper and we only demonstrate the performance of our GAN setup by measuring the likelihood fitness and machine learning efficacy metrics \cite{xu2019modeling}. The general overview of our proposed BCT-GAN is shown in Fig.~\ref{fig:overview-ECT}.

\subsection{Data Preprocessing: Continuous Column Normalization}
\label{sec:preprocess}
Tabular databases contain continuous, binary (boolean), and discrete columns with variable lengths. Therefore, they should be correctly represented and meaningfully distributed before using them for training purposes. Representing boolean and discrete fields are straightforward since they can be transformed into a block of one-hot floating-point vectors adjusted in the range of $\left [ -1,1 \right ]$ using the conventional $\tanh$ function \cite{choi2017generating}. Conversely, continuous columns are often non-Gaussian and resemble a multimodal distribution. This obliges to implement a fundamental preprocessing operation for transforming each column of the dataset according to a ordinal policy. The policy which we use herein is based on the variational Gaussian mixture model (VGM) \cite{bishop2007pattern} since its superior performance has been demonstrated for a variety of comprehensive tabular datasets \cite{xu2019modeling}. Inspired from CT-GAN, we also fit a VGM with $\varrho$ modes on every continuous column $C_{i}$ of the given dataset but we impose the following straightforward probability fairness condition \cite{hogg2005introduction}:
\begin{equation}
    p_{C_{i}}(c_{i,j})=\sum_{\tau=1}^{\varrho}\hat{\mu}_{\tau} \bigg( \mathcal{N}\Big(c_{i,j}\mid \mu_{\tau},\sigma_{\tau}^{2}\Big) \bigg), \quad \hat{\mu}_{\tau} \sim  \mathcal{N}(0,sI)
    \label{eq:probuniformity}
\end{equation}
\noindent where $c_{i,j}$ denotes the $j$-th field of $C_{i}$. For supporting the smoothness in computing the probability density function of every mode (i.e., every single Gaussian model of the VGM) in the mixture setup (see a relevant discussion in \cite{nualart2006malliavin}), we empirically set $s \rightarrow 0.5$. Statistically, Eq.~\ref{eq:probuniformity} centralizes all the modes, around the most dominant probability distribution with measurable displacements and avoids discarding others with short skewness \cite{friedman2017elements}. Herein, $\mu_{\tau}$ and $\sigma_{\tau}$ indicate the mean and standard deviation parameters of every mode in the actual VGM model. Our proposed condition in Eq.~\ref{eq:probuniformity} acts fairly since the term $\hat{\mu}_{\tau}$ is employed to resist against marginalizing Gaussian models with potentially trivial $\mu_{\tau}$ and $\sigma_{\tau}$ parameters (in terms of resembling low entropy \cite{hogg2005introduction,frenken2007entropy}). The accuracy of this action is highly dependent to the domain of $s$ and $\varrho$. For instance, setting $s \rightarrow 0$ might negatively affect the entire VGM with the possibility of eliminating non-trivial modes \cite{dineen2005non}.  

To the best of our knowledge, there is no analytical approach for finding the optimal value for $\varrho$ and it should be approximated according to the properties of the training dataset. Finally, we transform every record ($\mathbf{r}_{i}$) of the dataset into a one-hot vector of probabilities coming from the achieved VGM, linearly appended with the vector of normalized discrete values as follows \cite{xu2019modeling}:
\begin{equation}
    \mathbf{r}_{i} \equiv  \left \langle \vec{\Omega}_{\mathrm{cont}_{i}}\oplus \vec{\Omega}_{\mathrm{disc}_{i}} \right \rangle, \quad \forall i \in \mathbb{N}^{\left | \mathrm{cont} \right |}\cdot \mathbb{N}^{\left | \mathrm{disc} \right |}
    \label{eq:records}
\end{equation}
\noindent where $\vec{\Omega}_{\mathrm{cont}_{i}}$ and $\vec{\Omega}_{\mathrm{disc}_{i}}$ denote the vectors of preprocessed continuous and discrete columns, respectively. Moreover, $\oplus$ is a mathematical symbol for appending operation (element-wise) \cite{cohn1981universal}. This operation provides a solid representation for successfully training a GAN \cite{xu2019modeling}. The major challenge of this preprocessing transformation is marginalizing the less frequent discrete columns while $\vec{\Omega}_{\mathrm{disc}_{i}}$ are in minority compared to $\vec{\Omega}_{\mathrm{cont}_{i}}$ \cite{mclachlan1988mixture}. For resolving such an issue, we can either weight the discrete components of $\mathbf{r}_{i}$ using the Cauchy operation (probability distribution function) or redistributing $p_{C_{i}}(\cdot)$ with smaller $\varrho$ in the VGM model (Eq.~\ref{eq:probuniformity}) \cite{ming2003background}.    

\subsection{Defining the Conditional term for the Generator}
The motivation behind employing the class-conditional platform over regular GANs is twofold. Firstly, it has been shown that the conditional platforms demonstrates a relatively higher performance over the conventional configurations (e.g., vanilla GAN) \cite{brock2018large}. Secondly, tabular datasets are still heterogeneous even after running the aforementioned preprocessing and one-hot vectorization procedures (Section~\ref{sec:preprocess}). However, the performance of a conditional GAN is highly dependent to correctly defining the conditional term. We propose to bind the generator network according to a non-Gaussian probability distribution with asymmetric density function and wider variance. This operation is designed to expand the learning domain of the generator without biasing it toward any columns $C_{i}$ \cite{westfall2013understanding}. Hence, we empirically opt for the $\chi^{2}_{\beta}$ distribution with $\beta$ degree of freedom (this is a hyperparameter) and define the following binary mask function \cite{gupta2020fundamentals}:
\begin{equation}
     \varpi_{i}(\alpha_{i}) =  \left\{\begin{matrix}
1 & \mathrm{if}\quad \alpha_{i} <\frac{1}{2} \\ 
0 & \mathrm{otherwise.}
\end{matrix}\right. \quad  \mathrm{and} \quad  \alpha_{i} \in p_{\alpha} \sim \chi^{2}_{\beta} \quad
\label{eq:maskmask}
\end{equation}
\noindent where $\alpha_{i}$ is a conditioning parameter randomly drawn from a predefined probability distribution $p_{\alpha}$ and $\varpi_{i}(\alpha)$ refers to \textit{the moment-generating function} of $\chi^{2}_{\beta}$ \cite{casella1990statistical}. Given the fact that, this function is not symmetric, imposing the condition of $\alpha_{i} <0.5$ in Eq.~\ref{eq:maskmask} smoothly increases the chance of yielding sparse vector for $\varpi_{i}(\alpha_{i})$. Not only this contributes to improving the computational complexity of the entire GAN model during training, but also forces the generator to be slightly more independent to the conditional vector (the prior information). Finally, we define our novel conditional term for the generator network as follows.
\begin{equation}
    \oplus_{i=1}^{n_{t}} := \varpi_{1}(\alpha_{1}) \oplus \varpi_{2}(\alpha_{2}) \oplus \cdots \oplus  \varpi_{n_{t}}(\alpha_{n_{t}})
    \label{eq:compoundMask}
\end{equation}
\noindent where $n_{t}$ refers to the total number of discrete fields in every record (see Eq.~\ref{eq:records}). In fact, this term is a compound mask function \cite{xu2019modeling} for $\vec{\Omega}_{\mathrm{disc}_{i}}$ which binds $\mathbf{r}_{i}$ to discrete vectors. In other words, Eq.~\ref{eq:compoundMask} selects which discrete fields should be shown to the generator network during training aiming at avoiding memorizing the order of original records. The motivation behind this binding is the simplicity of conducting numerical operations on these vectors compared to the Gaussian models of the VGM. Moreover, similar to the CT-GAN, it enables training-by-sampling procedure \cite{xu2019modeling}. In the following subsection, we provide more details about architecture of the generator.

\subsection{Designed Architecture for the BCT-GAN}
As shown in Fig.~\ref{fig:overview-ECT}, our proposed BCT-GAN employs one generator and two independent discriminator networks. In one hand, such configuration relatively improves stability of the entire model since ideally, multiple discriminators should provide more gradients to the generator and, to some extent, avoid extreme instabilities, especially at early iterations \cite{hardy2019md}. On the other hand, exploiting multiple discriminators dramatically increases the total number of training parameters. Thus, we opt for two discriminators in order to circumvent such a potential side-effect and obtain a reasonable balance.

The designed architecture for the generator is a fully convolutional deep neural network (DNN) with five hidden layers. The input layer deploys $\mathbf{z}_{i} \in \mathbb{R}^{\left | \mathbf{r}_{i} \right |}$ where $\left | \mathbf{r}_{i} \right |$ denotes the total length of the discrete and continuous fields in every record. The first hidden layer implements $256$ filters padded with $5\times 5$ receptive field and $8 \times 32$ channels followed by batch normalization and ReLU activation function. Subsequent layers exploit $512$ filters plus $16\times 64$ channels, skip-$z$ \cite{brock2018large}, weight normalization \cite{salimans2016weight} and $\tanh$ activation function. Finally, the output layer exploits a fully connected layer with transposed convolution \cite{mao2018effectiveness} and gumbel softmax (with ratio $0.2$) \cite{jang2016categorical}. Furthermore, we implement the chordal distance minimization operation \cite{esmaeilpour2020class} to partially avoid potential extreme instabilities.

For simplicity and avoiding unnecessary complication during training, we implement an identical architecture for both discriminator networks. This unique architecture is also fully convolutional and requires a vector with dimension $\left | \mathbf{r}_{i} \right |$ in the input layer. There are three hidden layers with $256$, $512$, and $1024$ filters followed by skip-$z$, weight normalization, and leaky ReLU activation function. All these layers are distributed over $16\times 64$ channels without dropout. The output layer is fully connected which yields a logit vector for updating the generator weights according to the training policy of the least-square GAN \cite{hong2019generative}.

\begin{table*}[h]
\centering
\scriptsize
\caption{Comparison of the GANs on four tabular datasets. Herein, $\mathcal{L}_{val}$ and $\mathcal{L}_{test}$ measure the correlation between each synthesized dataset and its associated ground-truth oracle \cite{xu2019modeling} during the 5-fold cross validation and test phases, respectively.}
\rotatebox{90}{
\begin{tabular}{ccccccccccccc}
\hline
\multicolumn{1}{c||}{\multirow{2}{*}{Method}} & \multicolumn{3}{c||}{Adult}                                                                                          & \multicolumn{3}{c||}{Census}                                                                                         & \multicolumn{3}{c||}{Credit}                                                                                         & \multicolumn{3}{c}{News}                                                                                            \\ \cline{2-13} 
\multicolumn{1}{c||}{}                        & \multicolumn{1}{c|}{$\mathcal{L}_{val}$} & \multicolumn{1}{c|}{$\mathcal{L}_{test}$} & \multicolumn{1}{c||}{$F_{1}$} & \multicolumn{1}{c|}{$\mathcal{L}_{val}$} & \multicolumn{1}{c|}{$\mathcal{L}_{test}$} & \multicolumn{1}{c||}{$F_{1}$} & \multicolumn{1}{c|}{$\mathcal{L}_{val}$} & \multicolumn{1}{c|}{$\mathcal{L}_{test}$} & \multicolumn{1}{c||}{$F_{1}$} & \multicolumn{1}{c|}{$\mathcal{L}_{val}$} & \multicolumn{1}{c|}{$\mathcal{L}_{test}$} & \multicolumn{1}{c}{$R^{2}$}  \\ \hline \hline
\multicolumn{1}{l||}{Ground-truth}                & \multicolumn{1}{c|}{$-$}            & \multicolumn{1}{c|}{$-$}             & \multicolumn{1}{c||}{$0.667$} & \multicolumn{1}{c|}{$-$}            & \multicolumn{1}{c|}{$-$}             & \multicolumn{1}{c||}{$0.486$} & \multicolumn{1}{c|}{$-$}            & \multicolumn{1}{c|}{$-$}             & \multicolumn{1}{c||}{$0.741$} & \multicolumn{1}{c|}{$-$}            & \multicolumn{1}{c|}{$-$}             & \multicolumn{1}{r}{$0.156$}  \\ \hline \hline
\multicolumn{1}{l||}{CLBN \cite{chow1968approximating}}                    & \multicolumn{1}{c|}{$-2.763$}            & \multicolumn{1}{c|}{$-3.184$}             & \multicolumn{1}{c||}{$0.341$} & \multicolumn{1}{c|}{$-5.172$}            & \multicolumn{1}{r|}{$-6.970$}             & \multicolumn{1}{c||}{$0.308$} & \multicolumn{1}{c|}{$-7.502$}            & \multicolumn{1}{c|}{$-8.106$}             & \multicolumn{1}{c||}{$0.411$} & \multicolumn{1}{c|}{$-5.129$}            & \multicolumn{1}{c|}{$-8.081$}             & \multicolumn{1}{c}{$-6.607$} \\ \hline
\multicolumn{1}{l||}{PrivBN \cite{zhang2017privbayes}}                  & \multicolumn{1}{c|}{$-2.631$}            & \multicolumn{1}{c|}{$-4.152$}             & \multicolumn{1}{c||}{$0.416$} & \multicolumn{1}{c|}{$-4.525$}            & \multicolumn{1}{c|}{$-5.149$}             & \multicolumn{1}{c||}{$0.109$} & \multicolumn{1}{c|}{$-6.317$}            & \multicolumn{1}{c|}{$-6.942$}             & \multicolumn{1}{c||}{$0.189$} & \multicolumn{1}{c|}{$-4.306$}            & \multicolumn{1}{c|}{$-6.449$}             & \multicolumn{1}{r}{$-4.710$} \\ \hline
\multicolumn{1}{l||}{MedGAN \cite{choi2017generating}}                  & \multicolumn{1}{c|}{$-2.985$}            & \multicolumn{1}{c|}{$-3.126$}             & \multicolumn{1}{c||}{$0.351$} & \multicolumn{1}{c|}{$-4.711$}            & \multicolumn{1}{c|}{$-5.542$}             & \multicolumn{1}{c||}{$0.011$} & \multicolumn{1}{c|}{$-5.820$}            & \multicolumn{1}{c|}{$-6.171$}             & \multicolumn{1}{c||}{$0.027$} & \multicolumn{1}{c|}{$-4.451$}            & \multicolumn{1}{c|}{$-5.993$}             & \multicolumn{1}{r}{$-8.632$} \\ \hline
\multicolumn{1}{l||}{Table-GAN \cite{park2018data}}               & \multicolumn{1}{c|}{$-3.899$}            & \multicolumn{1}{c|}{$-5.270$}             & \multicolumn{1}{c||}{$0.511$} & \multicolumn{1}{c|}{$-4.753$}            & \multicolumn{1}{c|}{$-5.381$}             & \multicolumn{1}{c||}{$0.154$} & \multicolumn{1}{c|}{$-5.776$}            & \multicolumn{1}{c|}{$-5.992$}             & \multicolumn{1}{c||}{$0.036$} & \multicolumn{1}{c|}{$-3.525$}            & \multicolumn{1}{c|}{$-4.766$}             & \multicolumn{1}{c}{$-3.663$} \\ \hline
\multicolumn{1}{l||}{MT-GAN \cite{mottini2018airline}}                  & \multicolumn{1}{c|}{$-3.704$}            & \multicolumn{1}{c|}{$-4.086$}             & \multicolumn{1}{c||}{$0.432$} & \multicolumn{1}{c|}{$-4.661$}            & \multicolumn{1}{c|}{$-4.748$}             & \multicolumn{1}{c||}{$0.163$} & \multicolumn{1}{c|}{$-5.106$}            & \multicolumn{1}{c|}{$-6.355$}             & \multicolumn{1}{c||}{$0.022$} & \multicolumn{1}{c|}{$-3.235$}            & \multicolumn{1}{c|}{$ -5.633$}            & \multicolumn{1}{r}{$-4.239$} \\ \hline
\multicolumn{1}{l||}{ITS-GAN \cite{chen2019faketables}}                 & \multicolumn{1}{c|}{$-3.393$}            & \multicolumn{1}{c|}{$-4.001$}             & \multicolumn{1}{c||}{$0.446$} & \multicolumn{1}{c|}{$-4.670$}            & \multicolumn{1}{c|}{$-5.118$}             & \multicolumn{1}{c||}{$0.172$} & \multicolumn{1}{c|}{$-5.251$}            & \multicolumn{1}{c|}{$-5.995$}             & \multicolumn{1}{c||}{$0.012$} & \multicolumn{1}{c|}{$-3.289$}            & \multicolumn{1}{c|}{$-4.488$}             & \multicolumn{1}{r}{$-5.415$} \\ \hline
\multicolumn{1}{l||}{CT-GAN \cite{xu2019modeling}}                  & \multicolumn{1}{c|}{$-2.442$}            & \multicolumn{1}{c|}{$-2.968$}             & \multicolumn{1}{c||}{$0.592$} & \multicolumn{1}{c|}{$-4.119$}            & \multicolumn{1}{c|}{$-4.615$}             & \multicolumn{1}{c||}{$0.387$} & \multicolumn{1}{c|}{$\mathbf{-4.248}$}            & \multicolumn{1}{c|}{$\mathbf{-4.703}$}             & \multicolumn{1}{c||}{$\mathbf{0.642}$} & \multicolumn{1}{c|}{$-3.614$}            & \multicolumn{1}{c|}{$-4.210$}             & \multicolumn{1}{r}{$0.018$} \\ \hline
\multicolumn{1}{l||}{BCT-GAN (ours)}                 & \multicolumn{1}{c|}{$\mathbf{-2.197}$}            & \multicolumn{1}{c|}{$\mathbf{-2.622}$}             & \multicolumn{1}{c||}{$\mathbf{0.618}$} & \multicolumn{1}{c|}{$\mathbf{-3.722}$}            & \multicolumn{1}{c|}{$\mathbf{-3.805}$}             & \multicolumn{1}{c||}{$\mathbf{0.419}$} & \multicolumn{1}{c|}{$-4.379$}            & \multicolumn{1}{c|}{$-4.897$}             & \multicolumn{1}{c||}{$0.619$} & \multicolumn{1}{c|}{$\mathbf{-3.142}$}            & \multicolumn{1}{c|}{$\mathbf{-3.589}$}             & \multicolumn{1}{r}{$\mathbf{0.131}$} \\ \hline
\multicolumn{13}{c}{Statistically, $R^{2}$ might result in negative values for underfitted regression models or those not capable of achieving accurate decision boundaries \cite{hogg2005introduction}.}                                                             
\label{table:results}
\end{tabular}
}
\end{table*}

\section{Experiments}
\label{sec:experiment}
This section provides the details of our conducted experiments on four public datasets which have been benchmarked for tabular data processing, particularly for synthesis purposes \cite{xu2019modeling}. Adult, Census, and News are among the selected datasets from the UCI online repository \cite{dua2017uci} and they contain extensive combinations of continuous, binary, and complex discrete records. Following the baseline approaches \cite{xu2019modeling}, we have also carried out some experiments on the popular Credit dataset taken from the \href{https://www.kaggle.com/mlg-ulb/creditcardfraud}{Kaggle machine learning archive.}

These four datasets not only are independent and different in terms of statistical distribution and latent properties, but they also have various number of continuous, boolean, and discrete columns in several formats. For instance, the average number of continuous columns in the Credit dataset is 29 while this number for Adult, Census, and News is 6, 7, and 45, respectively. Moreover, the dataset with the highest number of discrete columns is Census (with 31 features) while the News dataset does not contain any discrete column. Another difference among these datasets is their benchmarking capacity. Specifically, three datasets as of Adult, Census, and Credit are benchmarked for classification tasks while News fits in the regression category. This diversity helps to effectively evaluate the performance of the generative models in different scales and capacities.

For training the GANs, we firstly transform and normalize all the records of the aforementioned datasets into the designated format as mentioned in Eq.~\ref{eq:records}. Toward this end, we empirically set $\varrho \rightarrow 10$ during fitting the VGM on the preprocessed datasets and fill up the potential empty fields with the null value. For fairness in comparison we constantly use Adam optimizer \cite{kingma2014adam} with $\left [ 0, 0.9 \right ]$ parameters and $1.8\cdot 10^{-3}$ learning rate. Similar to \cite{brock2018large}, we also implement an exploratory operation for tuning the optimal number of steps for training the generators over discriminators. Since our proposed synthesis approach employs two discriminators, we opt for two and three steps over $D_{1}(\cdot)$ and $D_{2}(\cdot)$, respectively. For other generative models we exploit two steps with the constant decay rate of $0.99$ on seven NVIDIA GTX-1080-Ti GPU with $9\times 11$ memory. Moreover, we use the static batch size of 500 with orthogonal regularization \cite{SaxeMG13} and spectral normalization \cite{miyato2018spectral} for both the generator and discriminator networks. We stop the training procedure as early as observing signs of instability or mode collapse \cite{brock2018large} (in average around 400 epochs). Eventually, the generator should craft a dataset with dimensions identical to the associated ground-truth oracle. For evaluation purposes, we organize the generated samples into two subsets of validation and test with ratio of 0.7 and 0.3, respectively.

For evaluating the performance of the generative models we measure two metrics. Firstly, we compute the likelihood fitness metric ($\mathcal{L}$) \cite{xu2019modeling} which measures the relative correlation ($\mathrm{RC}$) of the synthesized and ground-truth datasets as the following \cite{xu2019modeling,eghbal2017likelihood}.
\begin{equation}
    \mathcal{L} \propto  \left \lfloor  \mathrm{RC} \Big(p_{g}(\mathbf{z}_{i}; \theta_{g}),p_{r}( \mathbf{r}_{i})\Big) \right \rfloor, \quad  \forall i \in \left \{ 1,2,\cdots, \left | \mathbf{r}_{i} \right | \right \}
\end{equation}
\noindent where $p_{g}$ denotes the generator's distribution. We compute this metric separately for the validation and test subsets denoted respectively by $\mathcal{L}_{val}$ and $\mathcal{L}_{test}$. Results of this experiment on four benchmarking datasets are summarized in Table~\ref{table:results}.

The second evaluation metric is about the benchmarking capacity which is also known as the machine learning efficacy \cite{xu2019modeling}. In other words, we compare the classification or regression performance of some front-end algorithms on the ground-truth and the synthesized datasets. Technically, the performance of such algorithms on a synthesized dataset should be very close to the associated ground-truth oracle. Thus for such a comparison, we employ $F_{1}$ and $R^{2}$ scores for the classification and regression tasks, respectively \cite{xu2019modeling}. Regarding the choice of the front-end algorithms, we follow the pipeline suggested by Xu {\it et al.}~\cite{xu2019modeling}. Specifically, we implement Adaboost (with 50 estimators), decision tree (with depth 30), DNN (with 40 hidden layers) for Adult, Census, and Credit datasets. Additionally, we fit linear regression and DNN (with 100 hidden layers) for the News dataset. Average $F_{1}$ and $R^{2}$ scores over these front-end algorithms are shown in Table~\ref{table:results}. As shown in this table, for three datasets, our proposed BCT-GAN outperforms other generative models as it achieves higher $\mathcal{L}_{val}$, $\mathcal{L}_{test}$, $F_{1}$, and $R^{2}$ values. However, for the Credit dataset, our approach competitively loses against CT-GAN. We conjecture that this is due to the dependency of our BCT-GAN to discrete columns for yielding $\mathbf{r}_{i}$s (see Eq.~\ref{eq:records}). Since the Credit dataset is entirely continuous and binary, hence this negatively affects the performance of our proposed GAN. We are determined to resolve this issue via separating discrete and continuous embeddings of Eq.~\ref{eq:records} in our future works.

\section{Conclusion}
In this paper, we introduced a bi-discriminator GAN for tabular data synthesis. The major novelties of our approach is firstly the development of a preprocessing scheme and secondly defining a solid conditional term for the generator network to improve the entire performance of the generative model. This term is a vector based on a masked function using $\chi^{2}_{\beta}$ probability density function for more effectively constraining over the generator and consequently better capturing the input sample distributions. We experimentally demonstrated that, for the majority of the cases, our proposed BCT-GAN outperforms the state-of-the-art approaches both in terms of likelihood fitness metric and machine learning efficacy.

\section*{Acknowledgment}
This work was funded by Fédération des Caisses Desjardins du Québec and Mitacs accelerate program with agreement number IT25105.

\bibliography{mybibfile}

\begin{thebibliography}{10}
\expandafter\ifx\csname url\endcsname\relax
  \def\url#1{\texttt{#1}}\fi
\expandafter\ifx\csname urlprefix\endcsname\relax\def\urlprefix{URL }\fi
\expandafter\ifx\csname href\endcsname\relax
  \def\href#1#2{#2} \def\path#1{#1}\fi

\bibitem{even2007economics}
A.~Even, G.~Shankaranarayanan, P.~D. Berger, Economics-driven data management:
  An application to the design of tabular data sets, IEEE Transactions on
  Knowledge and Data Engineering 19~(6) (2007) 818--831.

\bibitem{shwartz2021tabular}
R.~Shwartz-Ziv, A.~Armon, Tabular data: Deep learning is not all you need,
  arXiv preprint arXiv:2106.03253.

\bibitem{clements2020sequential}
J.~M. Clements, D.~Xu, N.~Yousefi, D.~Efimov, Sequential deep learning for
  credit risk monitoring with tabular financial data, arXiv preprint
  arXiv:2012.15330.

\bibitem{buczak2015survey}
A.~L. Buczak, E.~Guven, A survey of data mining and machine learning methods
  for cyber security intrusion detection, IEEE Communications surveys \&
  tutorials 18~(2) (2015) 1153--1176.

\bibitem{ulmer2020trust}
D.~Ulmer, L.~Meijerink, G.~Cin{\`a}, Trust issues: Uncertainty estimation does
  not enable reliable ood detection on medical tabular data, in: Machine
  Learning for Health, PMLR, 2020, pp. 341--354.

\bibitem{xu2020synthesizing}
L.~Xu, et~al., Synthesizing tabular data using conditional gan, Ph.D. thesis,
  Massachusetts Institute of Technology (2020).

\bibitem{aven2010risk}
T.~Aven, O.~Renn, Risk management, in: Risk Management and Governance,
  Springer, 2010, pp. 121--158.

\bibitem{kornfeld1998automatically}
W.~Kornfeld, J.~Wattecamps, Automatically locating, extracting and analyzing
  tabular data, in: Proceedings of the 21st annual international ACM SIGIR
  conference on Research and development in information retrieval, 1998, pp.
  347--348.

\bibitem{cartella2021adversarial}
F.~Cartella, O.~Anunciacao, Y.~Funabiki, D.~Yamaguchi, T.~Akishita,
  O.~Elshocht, Adversarial attacks for tabular data: Application to fraud
  detection and imbalanced data, arXiv preprint arXiv:2101.08030.

\bibitem{sheth1990federated}
A.~P. Sheth, J.~A. Larson, Federated database systems for managing distributed,
  heterogeneous, and autonomous databases, ACM Computing Surveys (CSUR) 22~(3)
  (1990) 183--236.

\bibitem{wang1990polygen}
Y.~R. Wang, S.~E. Madnick, et~al., A polygen model for heterogeneous database
  systems: The source tagging perspective.

\bibitem{borisov2021deep}
V.~Borisov, T.~Leemann, K.~Se{\ss}ler, J.~Haug, M.~Pawelczyk, G.~Kasneci, Deep
  neural networks and tabular data: A survey, arXiv preprint arXiv:2110.01889.

\bibitem{loorak2016exploring}
M.~H. Loorak, C.~Perin, C.~Collins, S.~Carpendale, Exploring the possibilities
  of embedding heterogeneous data attributes in familiar visualizations, IEEE
  Transactions on Visualization and Computer Graphics 23~(1) (2016) 581--590.

\bibitem{khan2020toward}
M.~A. Khan, J.~Kim, Toward developing efficient conv-ae-based intrusion
  detection system using heterogeneous dataset, Electronics 9~(11) (2020) 1771.

\bibitem{socher2012deep}
R.~Socher, Y.~Bengio, C.~D. Manning, Deep learning for nlp (without magic), in:
  Tutorial Abstracts of ACL 2012, 2012, pp. 5--5.

\bibitem{traquair2019deep}
M.~Traquair, E.~Kara, B.~Kantarci, S.~Khan, Deep learning for the detection of
  tabular information from electronic component datasheets, in: 2019 IEEE
  Symposium on Computers and Communications (ISCC), IEEE, 2019, pp. 1--6.

\bibitem{gorishniy2021revisiting}
Y.~Gorishniy, I.~Rubachev, V.~Khrulkov, A.~Babenko, Revisiting deep learning
  models for tabular data, arXiv preprint arXiv:2106.11959.

\bibitem{bourou2021review}
S.~Bourou, A.~El~Saer, T.-H. Velivassaki, A.~Voulkidis, T.~Zahariadis, A review
  of tabular data synthesis using gans on an ids dataset, Information 12~(9)
  (2021) 375.

\bibitem{goodfellow2014generative}
I.~Goodfellow, J.~Pouget-Abadie, M.~Mirza, B.~Xu, D.~Warde-Farley, S.~Ozair,
  A.~Courville, Y.~Bengio, Generative adversarial nets, in: Adv Neural Inf
  Process Syst, 2014, pp. 2672--2680.

\bibitem{shanmugam2020and}
D.~Shanmugam, D.~Blalock, G.~Balakrishnan, J.~Guttag, When and why test-time
  augmentation works, arXiv preprint arXiv:2011.11156.

\bibitem{tsechansky1999mining}
M.~S. Tsechansky, N.~Pliskin, G.~Rabinowitz, A.~Porath, Mining relational
  patterns from multiple relational tables, Decision Support Systems 27~(1-2)
  (1999) 177--195.

\bibitem{shabtai2012survey}
A.~Shabtai, Y.~Elovici, L.~Rokach, A survey of data leakage detection and
  prevention solutions, Springer Science \& Business Media, 2012.

\bibitem{alneyadi2016survey}
S.~Alneyadi, E.~Sithirasenan, V.~Muthukkumarasamy, A survey on data leakage
  prevention systems, Journal of Network and Computer Applications 62 (2016)
  137--152.

\bibitem{kingma2013auto}
D.~P. Kingma, M.~Welling, Auto-encoding variational bayes, in: 2nd
  International Conference on Learning Representations, {ICLR} 2014, Banff, AB,
  Canada, April 14-16, 2014, Conference Track Proceedings, 2014.

\bibitem{chow1968approximating}
C.~Chow, C.~Liu, Approximating discrete probability distributions with
  dependence trees, IEEE transactions on Information Theory 14~(3) (1968)
  462--467.

\bibitem{zhang2017privbayes}
J.~Zhang, G.~Cormode, C.~M. Procopiuc, D.~Srivastava, X.~Xiao, Privbayes:
  Private data release via bayesian networks, ACM Transactions on Database
  Systems (TODS) 42~(4) (2017) 1--41.

\bibitem{ma2020vaem}
C.~Ma, S.~Tschiatschek, R.~E. Turner, J.~M. Hern{\'{a}}ndez{-}Lobato, C.~Zhang,
  {VAEM:} a deep generative model for heterogeneous mixed type data, in:
  Advances in Neural Information Processing Systems 33: Annual Conference on
  Neural Information Processing Systems 2020, NeurIPS 2020, December 6-12,
  2020, virtual, 2020.

\bibitem{park2018data}
N.~Park, M.~Mohammadi, K.~Gorde, S.~Jajodia, H.~Park, Y.~Kim, Data synthesis
  based on generative adversarial networks, Proc. {VLDB} Endow. 11~(10) (2018)
  1071--1083.
\newblock \href {http://dx.doi.org/10.14778/3231751.3231757}
  {\path{doi:10.14778/3231751.3231757}}.

\bibitem{genevay2017gan}
A.~Genevay, G.~Peyr{\'e}, M.~Cuturi, Gan and vae from an optimal transport
  point of view, arXiv preprint arXiv:1706.01807.

\bibitem{mi2018probe}
L.~Mi, M.~Shen, J.~Zhang, A probe towards understanding gan and vae models,
  arXiv preprint arXiv:1812.05676.

\bibitem{feizi2017understanding}
S.~Feizi, F.~Farnia, T.~Ginart, D.~Tse, Understanding gans in the {LQG}
  setting: Formulation, generalization and stability, {IEEE} J. Sel. Areas Inf.
  Theory 1~(1) (2020) 304--311.
\newblock \href {http://dx.doi.org/10.1109/jsait.2020.2991375}
  {\path{doi:10.1109/jsait.2020.2991375}}.

\bibitem{goodfellow2016deep}
I.~Goodfellow, Y.~Bengio, A.~Courville, Y.~Bengio, Deep learning, Vol.~1, MIT
  press Cambridge, 2016.

\bibitem{choi2017generating}
E.~Choi, S.~Biswal, B.~Malin, J.~Duke, W.~F. Stewart, J.~Sun, Generating
  multi-label discrete patient records using generative adversarial networks,
  in: Machine learning for healthcare conference, PMLR, 2017, pp. 286--305.

\bibitem{camino2018generating}
R.~Camino, C.~Hammerschmidt, R.~State, Generating multi-categorical samples
  with generative adversarial networks, arXiv preprint arXiv:1807.01202.

\bibitem{radford2015unsupervised}
A.~Radford, L.~Metz, S.~Chintala, Unsupervised representation learning with
  deep convolutional generative adversarial networks, in: 4th Intl Conf Learn
  Repres, 2016.

\bibitem{mottini2018airline}
A.~Mottini, A.~Lheritier, R.~Acuna-Agost, Airline passenger name record
  generation using generative adversarial networks, arXiv preprint
  arXiv:1807.06657.

\bibitem{mroueh2017sobolev}
Y.~Mroueh, C.~Li, T.~Sercu, A.~Raj, Y.~Cheng, Sobolev {GAN}, in: 6th Intl Conf
  Learn Repres, 2018.

\bibitem{bellemare2017cramer}
M.~G. Bellemare, I.~Danihelka, W.~Dabney, S.~Mohamed, B.~Lakshminarayanan,
  S.~Hoyer, R.~Munos, The cramer distance as a solution to biased wasserstein
  gradients, CoRR abs/1705.10743.

\bibitem{chen2019faketables}
H.~Chen, S.~Jajodia, J.~Liu, N.~Park, V.~Sokolov, V.~Subrahmanian, Faketables:
  Using gans to generate functional dependency preserving tables with bounded
  real data., in: IJCAI, 2019, pp. 2074--2080.

\bibitem{muller1997integral}
A.~M{\"u}ller, Integral probability metrics and their generating classes of
  functions, Adv in Applied Probability (1997) 429--443.

\bibitem{mroueh2017fisher}
Y.~Mroueh, T.~Sercu, Fisher {GAN}, in: Adv in Neural Inf Proc Sys 30: Annual
  Conf on Neural Inf Proc Sys, 2017, pp. 2513--2523.

\bibitem{kohavi1996scaling}
R.~Kohavi, et~al., Scaling up the accuracy of naive-bayes classifiers: A
  decision-tree hybrid., in: Kdd, Vol.~96, 1996, pp. 202--207.

\bibitem{xu2019modeling}
L.~Xu, M.~Skoularidou, A.~Cuesta{-}Infante, K.~Veeramachaneni, Modeling tabular
  data using conditional {GAN}, in: Advances in Neural Information Processing
  Systems 32: Annual Conference on Neural Information Processing Systems 2019,
  NeurIPS 2019, December 8-14, 2019, Vancouver, BC, Canada, 2019, pp.
  7333--7343.

\bibitem{bishop2007pattern}
C.~M. Bishop, \href{https://www.worldcat.org/oclc/71008143}{Pattern recognition
  and machine learning, 5th Edition}, Information science and statistics,
  Springer, 2007.
\newline\urlprefix\url{https://www.worldcat.org/oclc/71008143}

\bibitem{hogg2005introduction}
R.~V. Hogg, J.~McKean, A.~T. Craig, Introduction to mathematical statistics,
  Pearson Education, 2005.

\bibitem{nualart2006malliavin}
D.~Nualart, The Malliavin calculus and related topics, Vol. 1995, Springer,
  2006.

\bibitem{friedman2017elements}
J.~H. Friedman, The elements of statistical learning: Data mining, inference,
  and prediction, springer open, 2017.

\bibitem{frenken2007entropy}
K.~Frenken, et~al., Entropy statistics and information theory, Chapters.

\bibitem{dineen2005non}
P.~Dineen, G.~Rocha, P.~Coles, Non-random phases in non-trivial topologies,
  Monthly Notices of the Royal Astronomical Society 358~(4) (2005) 1285--1289.

\bibitem{cohn1981universal}
P.~M. Cohn, P.~M. Cohn, Universal algebra, Vol. 159, Reidel Dordrecht, 1981.

\bibitem{mclachlan1988mixture}
G.~J. McLachlan, K.~E. Basford, Mixture models: Inference and applications to
  clustering, Vol.~38, M. Dekker New York, 1988.

\bibitem{ming2003background}
Y.~Ming, J.~Jiang, J.~Ming, Background modeling and subtraction using a
  local-linear-dependence-based cauchy statistical model., in: DICTA, 2003, pp.
  469--478.

\bibitem{brock2018large}
A.~Brock, J.~Donahue, K.~Simonyan, Large scale {GAN} training for high fidelity
  natural image synthesis, in: Intl Conf Learn Repres, 2019.

\bibitem{westfall2013understanding}
P.~H. Westfall, K.~S. Henning, Understanding advanced statistical methods, CRC
  Press Boca Raton, FL, USA:, 2013.

\bibitem{gupta2020fundamentals}
S.~Gupta, V.~Kapoor, Fundamentals of mathematical statistics, Sultan Chand \&
  Sons, 2020.

\bibitem{casella1990statistical}
G.~Casella, R.~L. Berger, Statistical inference. wadsworth \& brooks, Cole,
  Pacific Grove, CA.

\bibitem{hardy2019md}
C.~Hardy, E.~Le~Merrer, B.~Sericola, Md-gan: Multi-discriminator generative
  adversarial networks for distributed datasets, in: 2019 IEEE international
  parallel and distributed processing symposium (IPDPS), IEEE, 2019, pp.
  866--877.

\bibitem{salimans2016weight}
T.~Salimans, D.~P. Kingma, Weight normalization: A simple reparameterization to
  accelerate training of deep neural networks, Advances in neural information
  processing systems 29 (2016) 901--909.

\bibitem{mao2018effectiveness}
X.~Mao, Q.~Li, H.~Xie, R.~Y. Lau, Z.~Wang, S.~P. Smolley, On the effectiveness
  of least squares generative adversarial networks, IEEE Trans on pattern
  analysis and machine intelligence 41~(12) (2018) 2947--2960.

\bibitem{jang2016categorical}
E.~Jang, S.~Gu, B.~Poole, Categorical reparameterization with gumbel-softmax,
  in: 5th International Conference on Learning Representations, {ICLR} 2017,
  Toulon, France, April 24-26, 2017, Conference Track Proceedings,
  OpenReview.net, 2017.

\bibitem{esmaeilpour2020class}
M.~Esmaeilpour, P.~Cardinal, A.~L. Koerich, Class-conditional defense gan
  against end-to-end speech attacks, in: IEEE Intl Conf Acoust, Speech and
  Signal Process, 2021, pp. 2565--2569.

\bibitem{hong2019generative}
Y.~Hong, U.~Hwang, J.~Yoo, S.~Yoon, How generative adversarial networks and
  their variants work: An overview, ACM Computing Surveys 52~(1) (2019) 1--43.

\bibitem{dua2017uci}
D.~Dua, C.~Graff, et~al., Uci machine learning repository.

\bibitem{kingma2014adam}
D.~P. Kingma, J.~Ba, Adam: {A} method for stochastic optimization, in: 3rd Intl
  Conf Learn Repres, 2015.

\bibitem{SaxeMG13}
A.~M. Saxe, J.~L. McClelland, S.~Ganguli, Exact solutions to the nonlinear
  dynamics of learning in deep linear neural networks, in: 2nd Intl Conf Learn
  Repres, 2014.

\bibitem{miyato2018spectral}
T.~Miyato, T.~Kataoka, M.~Koyama, Y.~Yoshida, Spectral normalization for
  generative adversarial networks, in: Intl Conf Learn Repres, 2018.

\bibitem{eghbal2017likelihood}
H.~Eghbal-zadeh, G.~Widmer, Likelihood estimation for generative adversarial
  networks, arXiv preprint arXiv:1707.07530.

\end{thebibliography}

\end{document}